\documentclass[10pt,twocolumn]{article}

\usepackage{titling}
\usepackage[noblocks]{authblk}

\usepackage{amsmath,amssymb}

\usepackage[french,english]{babel}
\usepackage[utf8]{inputenc}
\usepackage[T1]{fontenc}

\usepackage{subcaption}
\usepackage{stfloats}
\usepackage{graphicx}
\usepackage{xspace}
\usepackage{pdfpages}
\usepackage{wrapfig}
\usepackage{wasysym}
\usepackage{enumitem}
\usepackage{adjustbox}
\usepackage{changepage}
\usepackage{ragged2e}
\usepackage{tikz}
\usepackage{longtable}
\usepackage{changepage}
\usepackage{setspace}
\usepackage{hhline}
\usepackage{tabto}
\usepackage{float}
\usepackage{multirow}
\usepackage{makecell}
\usepackage{fancyhdr}
\usepackage[colorlinks=true,urlcolor=blue,linkcolor=black,citecolor=black]{hyperref}

\usepackage[disable,textsize=footnotesize]{todonotes}
\begin{document}

\title{The GeoLifeCLEF 2020 Dataset} 

\author[1]{Elijah Cole\thanks{equal contribution}\thanks{ecole@caltech.edu}}
\author[2,3,4]{Benjamin Deneu\thanksmark{1}\thanks{benjamin.deneu@inria.fr}}
\author[3,2]{Titouan Lorieul}
\author[3,5]{Maximilien Servajean} 
\author[6]{Christophe Botella}
\author[7]{Dan Morris}
\author[7]{Nebojsa Jojic}
\author[4]{Pierre Bonnet} 
\author[2,3]{Alexis Joly} 

\affil[1]{Caltech, Pasadena, US}
\affil[2]{Inria, Montpellier, France}
\affil[3]{LIRMM, CNRS, University of Montpellier, Montpellier, France}
\affil[4]{CIRAD, AMAP, Montpellier, France}
\affil[5]{AMIS, Paul Valery University - Montpellier 3, Montpellier, France}
\affil[6]{CNRS, LECA, Grenoble, France}
\affil[7]{Microsoft Research, Redmond, US}

\date{}

\maketitle
\vspace*{-2cm}

\abstract{ 
Understanding the geographic distribution of species is a key concern in conservation.
By pairing species occurrences with environmental features, researchers can model the relationship between an environment and the species which may be found there. 
To facilitate research in this area, we present the GeoLifeCLEF 2020 dataset, which consists of 1.9 million species observations paired with high-resolution remote sensing imagery, land cover data, and altitude, in addition to traditional low-resolution climate and soil variables.
We also discuss the GeoLifeCLEF 2020 competition, which aims to use this dataset to advance the state-of-the-art in location-based species recommendation.
}

\section{Introduction}

\begin{figure}[tb!]
    \centering
    \includegraphics[width=0.35\textwidth]{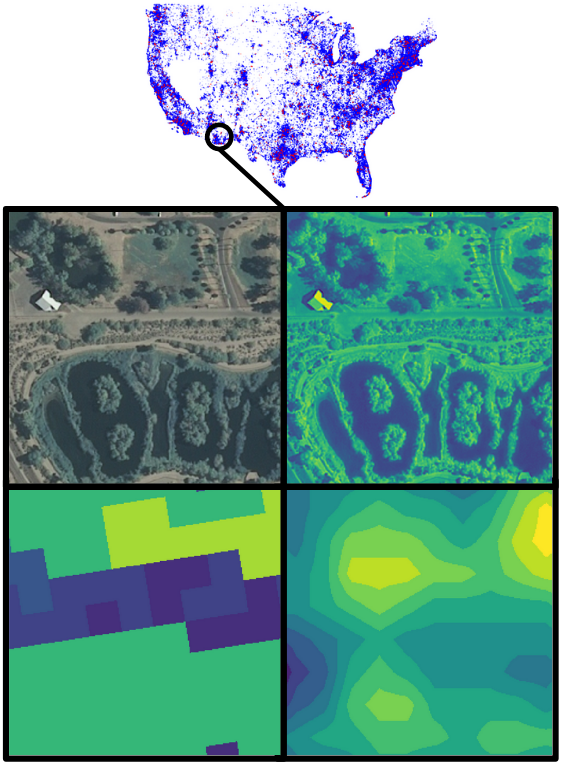}
    \caption{Each species observation is paired with high-resolution covariates (clockwise from top left: RGB imagery, IR imagery, altitude, land cover).}
    \label{fig:splash}
\end{figure}

In order to make informed conservation decisions it is essential to understand where different species live.
Citizen science projects now generate millions of geo-located species observations every year, covering tens of thousands of species. 
But how can these point observations be used to predict what species might be found at a new location? 

A common approach is to build a \emph{species distribution model} (SDM) \cite{elith2009sdm}, which uses a location's \emph{environmental covariates} (e.g. temperature, elevation, land cover) to predict which species may be found there.
Once trained, the model can be used to make predictions for any location where those covariates are available. 

Developing an SDM requires a dataset where each species occurrence is paired with a collection of environmental covariates.
However, many existing SDM datasets are both highly specialized and not readily accessible, having been assembled by scientists studying particular species or regions.
In addition, the provided environmental covariates are typically coarse, with resolutions ranging from hundreds of meters to kilometers per pixel. 

\begin{figure*}
    \centering
    \hfill
    \begin{subfigure}{.45\textwidth}
        \includegraphics[width=\textwidth]{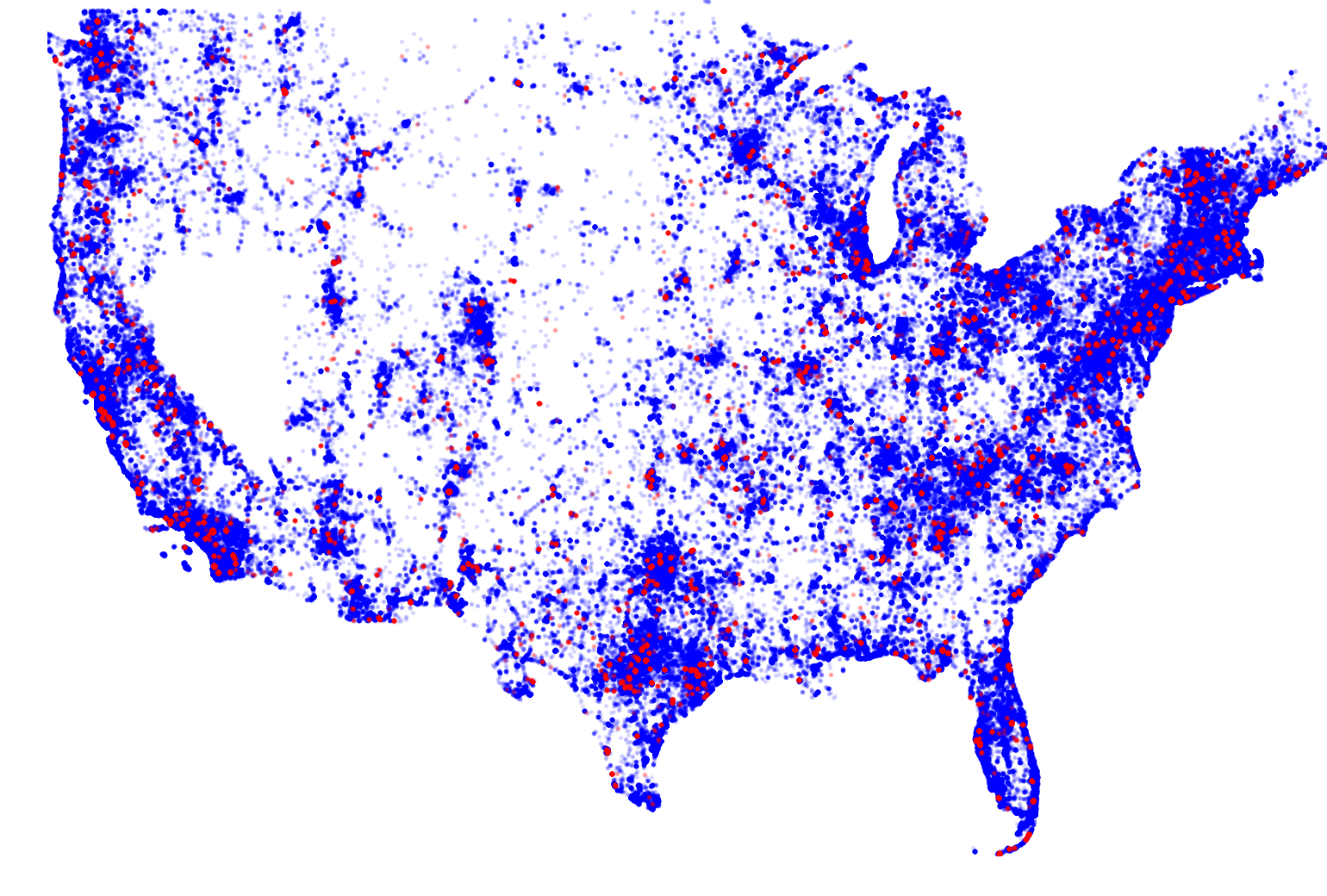}
        \caption{US}
    \end{subfigure}
    \hfill
    \begin{subfigure}{.45\textwidth}
        \includegraphics[width=\textwidth]{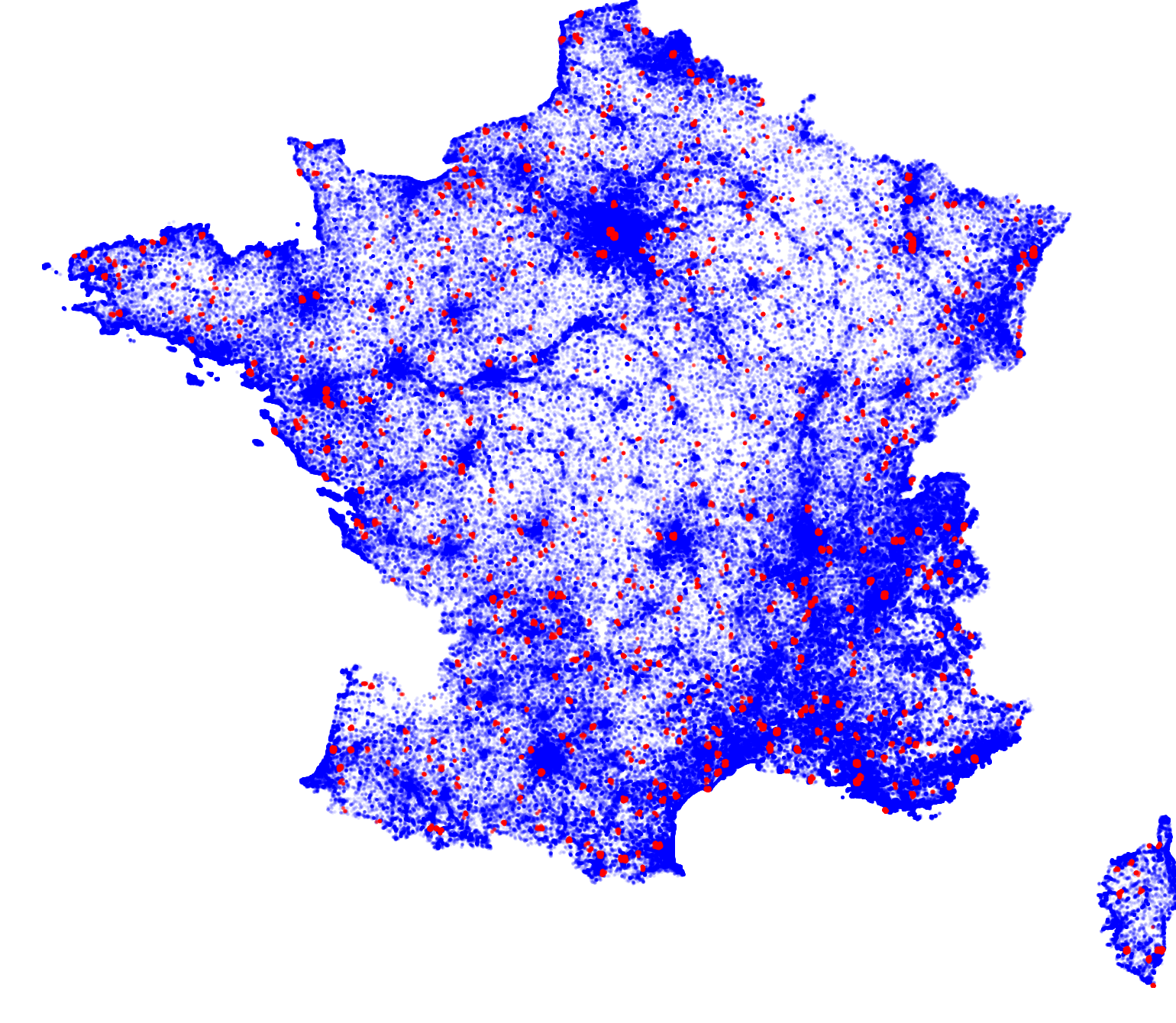}
        \caption{France}
    \end{subfigure}
    \hfill~
    \caption{
        Occurrences distribution over the US and France.
        Blue dots represent training data, red dots represent test data.
    }
    \label{fig:occurrences distribution}
\end{figure*}

In this work we present the GeoLifeCLEF 2020 dataset, which consists of over 1.9 million observations covering $31,435$ plant and animal species. 
Each species observation is paired with high-resolution covariates (RGB-IR imagery, land cover, and altitude) as illustrated in \autoref{fig:splash}.
These high-resolution covariates are resampled to a spatial resolution of 1 meter per pixel and provided as $256\times 256$ images covering a 256m $\times$ 256m square centered on each observation.
To the best of our knowledge, this is the first publicly available dataset to pair remote sensing imagery with species observations. 
We also provide traditional coarse covariates (climate and soil variables).
In total, the dataset is approximately 840GB -- see \autoref{tab:datasize} for more details.
Our hope is that this analysis-ready dataset will (i) make the SDM problem more accessible to machine learning researchers and (ii) facilitate novel research in large-scale, high-resolution and remote-sensing-based species distribution modeling. 

\section{Data Preparation}

\subsection{Occurrence Data}

\paragraph{US.}
Species occurrences for the US are collected from the citizen science project iNaturalist \cite{inatWeb}. 
In particular, the provided occurrences are derived from the \emph{iNaturalist Research-grade Observations} Dataset as it appeared on January 8, 2020 \cite{inatDownloadUS}.
To qualify as \emph{research-grade}, an iNaturalist observations must correspond to a wild organism (not captive or cultivated) and must be associated with a timestamp and date. 
In addition, we filter down to observations that (i) fall inside the contiguous United States, (ii) have a coordinate uncertainty of less than 30m, and (iii) are covered by valid remote sensing imagery.\footnote{We define ``valid'' remote sensing imagery to be 4-band (RGB-IR) imagery. This accounts for the absence of data points in Nevada, where only 3-band (RGB) imagery was available from the database we used.}
To further reduce the size of the dataset, we kept only observations from 2019. 
The result is $1,097,640$ occurrences for the US, covering $25,459$ species. 
In contrast to the data for France described below, there was no per-species subsampling nor were any species excluded based on the number of available observations. 

\paragraph{France.}
Species occurrences for France are drawn from two sources.
Plant occurrences are extracted from the citizen science project Pl@ntnet \cite{plantnetWeb}, which is the largest source of plant occurrences in France with a precise geo-localization.
We considered only those occurrences collected between January 2017 and October 2019.
These occurrences are filtered down to keep observations with (i) a confidence score (as predicted by the Pl@ntnet classification model) greater than $0.9$, (ii) a coordinate uncertainty lower than 30m, and (iii) available remote sensing imagery\footnote{In particular, observations in the department of Gironde (33) have been removed because remote sensing data was unavailable.}.
Species having only one observation are then removed, as well as species of \textit{Phalaenopsis} that are over-represented and not present in the natural environment.
For plant species with over $1,500$ occurrences, only $1,500$ randomly sampled observations are kept. 
Animal occurrences come from the \emph{iNaturalist Research-grade Observations} Dataset as it appeared on February 11, 2020 \cite{inatDownloadFrance}.
The result is $823,483$ occurrences for France, covering $8,075$ species.

\paragraph{Common taxonomy.}
Species names associated to US and French occurrences were aligned using the binomial name of the species (\textit{e.g.\ }``\textit{Acer campestre L.}'' is converted to ``\textit{Acer campestre}''). 
Observations from iNaturalist were already associated with binomial names and matched to the Global Biodiversity Information Facility (GBIF) taxonomy \cite{gbifweb}.
To avoid duplicating species, we used the GBIF API \cite{GBIFAPI} to assign binomial names to Pl@ntnet occurrence.
We found 2,099 species in common between US and French occurrences.

\subsection{Remote Sensing Imagery}

\paragraph{US.}
Observations in the US are paired with RGB-IR imagery from the 2009-2011 cycle of the National Agriculture Imagery Program (NAIP) \cite{naip}. 
For each observation, we find all digital ortho quarter quad (DOQQ) tiles covering that location. 
If an observation is covered by more than one DOQQ tile, we choose a tile that contains the entirety of the 256 m $\times$ 256 m square centered at that observation (so there will be no ``seams'' in the imagery).\footnote{If more than one tile meets this criterion, we choose at random.}
This is almost always achievable because of the substantial overlap between adjacent DOQQs. 

\paragraph{France.}
Observations in France are paired with RGB-IR imagery from BD ORTHO\up{\textregistered} 2.0 and ORTHO HR\up{\textregistered} 1.0 databases from the French National Institute of Geographic and Forest Information (IGN) \cite{BDOrtho}.
This imagery has a spatial resolution of 0.5 meters per pixel, but was resampled to 1 meter per pixel to match the US data. 

\subsection{Land Cover Data}

\paragraph{US.}
The US land cover data is from the 2011 edition of the National Land Cover Database (NLCD) \cite{nlcd2011}.
NLCD has a spatial resolution of 30 meters per pixel, which has been resampled to 1 meter per pixel without interpolation. 
The land cover classes are listed in \autoref{tab:landcoverclasses}.
Four NLCD classes (Dwarf Scrub, Sedge/Herbaceous, Lichens, Moss) appear only in Alaska and have therefore been omitted. 

\paragraph{France.}
The France land cover data is from the 2016 edition of the Centre d'Etudes Spatiales de la Biosphère (CESBIO) \cite{cesbiodata}.
CESBIO has a spatial resolution of 10 meters per pixel, which has been been re-sampled to 1 meter per pixel (no interpolation).
The land cover classes are listed in \autoref{tab:landcoverclasses}.

\paragraph{Suggested alignment.}
We list the land cover classes of both countries in \autoref{tab:landcoverclasses}.
This table includes the the class codes from the original datasets as well as the new class codes used in our dataset.
We also provide a suggested alignment between the two sets of land cover classes in \autoref{tab:landcoveralignment}.

\subsection{Elevation Data}
Elevation data comes from the NASA Shuttle Radar Topography Mission (SRTM), particularly the SRTMGL1 (v003) dataset \cite{srtm}.
Since this data has global coverage, we use it for both the US and France. 
The pixel resolution is 1 arc second.
The data was re-sampled to 1 meter per pixel by projecting arc seconds to meters in latitude-dependent local coordinates.
This conversion was done using a spherical model of the earth with a radius of 6,371 kilometers.
In this model, 1 arc second in latitude is constant and approximately equal to 30.87 meters.
For a given position with a latitude $lat$ in radians, the conversion from 1 arc second (arcsec) into meters is computed using:
\begin{align*}
    1\ \textrm{arcsec in latitude}\ &=\ 30.87\ \textrm{meters} \\
    1\ \textrm{arcsec in longitude}\ &=\ 30.87 \times \cos(lat)\ \textrm{meters}
\end{align*}
The error in this approximation is small - around 0.5 meters for an arc 256 meters long.\todo{at the latitudes we're considering? At what latitude was this calculation done?} 

\subsection{Bioclimatic and Pedologic Rasters}
We provide 19 bioclimatic and 8 pedologic (soil) variables which cover both the US and France.
These 27 variables are summarized in \autoref{tab:my_label}.
These variables are provided as rasters in the WGS84 coordinates system.

The bioclimatic variables come from WorldClim 1.4, a commonly used climatic database for SDM applications \cite{hijmans2005very}.
WorldClim variables are produced by statistical models which are based on measurements from many meteorological stations from around the world.
The spatial resolution is approximately 1 kilometer.
Pedologic variables are from SoilGrids \cite{hengl2017soilgrids250m} and represent physico-chemical soil properties such as soil depth, texture, pH, and organic matter content.
These values are based on machine learning models trained on around 150,000 soil profiles around the world and 158 remote-sensing based covariates.

\subsection{Train/Test Splitting Procedure}
\begin{figure}
    \centering
    \includegraphics[width=.4\textwidth]{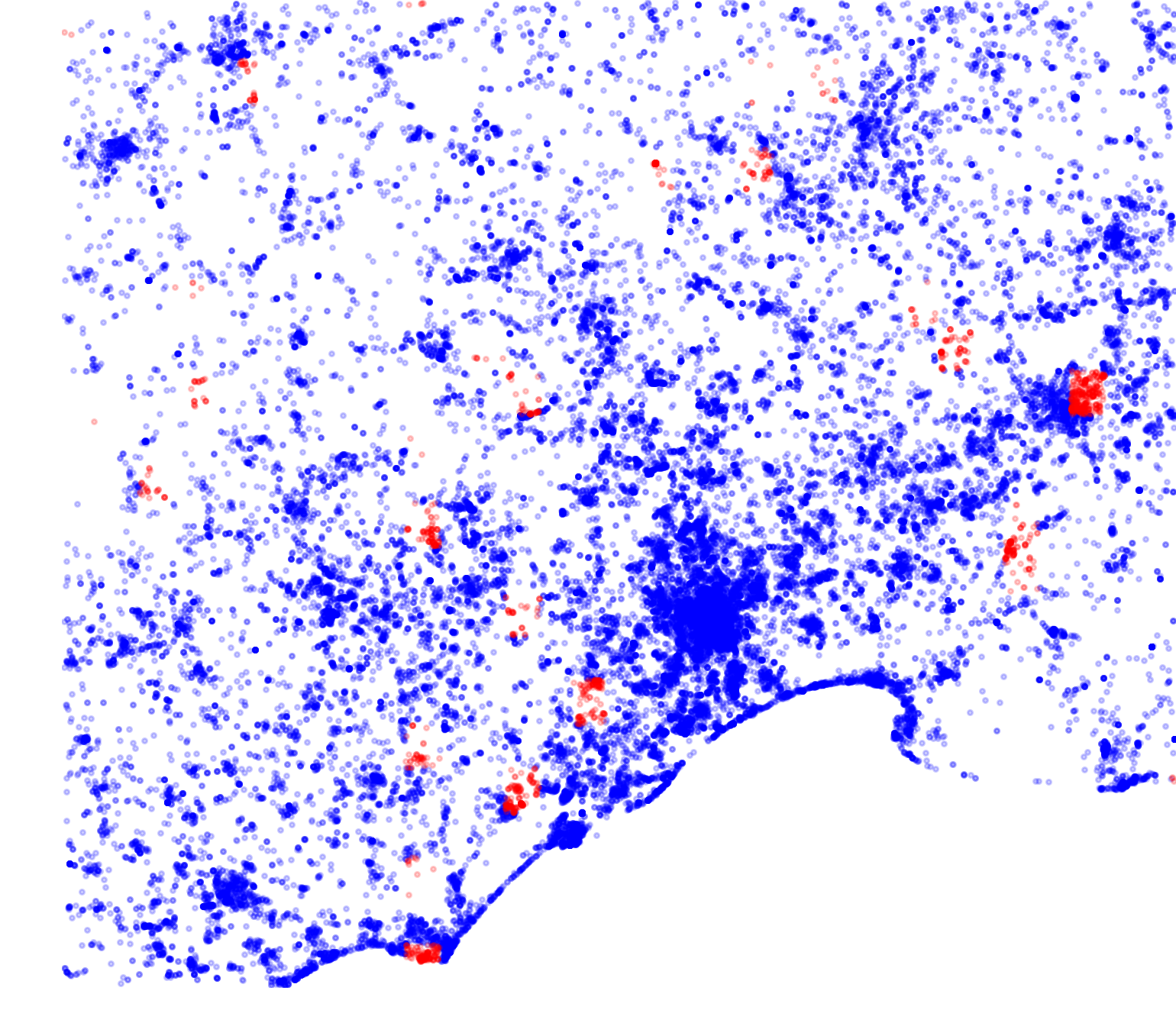}
    \caption{
        Close-up view of the occurrences distribution from the area around Montpellier, France, illustrating of the spatial block holdout procedure.
        Blue and red dots represent respectively the training and test data.
    }
    \label{fig:split-illustration}
\end{figure}

The full set of occurrences was split in a training and testing set using a spatial block holdout procedure.
This allows us to evaluate models in a way that limits the effect of the spatial bias in the data. 
We chose to base our split on a grid of 5 km $\times$ 5 km quadrats.
We randomly sampled 2.5\% of the quadrats for the test set, and assigned the remaining quadrats to the training set.
This procedure is illustrated in \autoref{fig:split-illustration}.
The test data (red) is drawn from different spatial blocks than the training data (blue).

\section{Evaluation}
Ground truth data for species presence and absence is extremely limited, primarily because it is very costly to verify species absences on a large scale.
The GeoLifeCLEF 2020 dataset consists of \emph{presence-only} data -- we know where species have been observed, but we do not know where they have been confirmed to be absent.
This precludes the use of standard multi-label classification metrics. 

We instead consider classification metrics such as the top-$K$ accuracy, which are well-defined in this setting.
However, this metric assumes that the number of species $K$ should be the same at each location.
In practice, different locations may be home to different numbers of species. 
Therefore the standard top-$K$ accuracy is used only as a secondary evaluation metric. 

The main evaluation criterion is an adaptive variant of the top-$K$ accuracy.
This metric is computed as follows. 
First, we make an inference pass over the data to produce a list of class (species) scores for each data point. 
We then calculate the threshold $t$ such that the average number of predicted species (i.e. species with a score greater than $t$) is $K$, where the average is computed over all data points. 
Note that each data point may have a different number of classes whose scores exceed $t$. 
Finally, we compute the percentage of test observations for which the correct species has a score above the threshold $t$.

More formally, if the species scores for the $n$-th observation are denoted $s_1^{(n)},s_2^{(n)},\dots,s_C^{(n)}$ where $C$ is the total number of species, then the average accuracy for a given value of threshold $t$ is computed using
\begin{equation*}
    \frac{1}{N} \sum_{n=1}^N \delta_{s_{y_n}^{(n)} \geq t}
\end{equation*}
where $y_n$ is the target species of the $n$-th sample and $\delta_Z$ is the indicator function on the statement $Z$ (equal to $1$ if $Z$ is true, $0$ otherwise).
Since there may not always exist a $t$ such that the average number of predicted species is \emph{exactly} $K$, we compute the threshold $t$ as
\begin{align*}
    t &= \min \left\{ t' \in \mathbb{R} \;\middle|\; \frac{1}{N} \sum_{n=1}^N \sum_{i=1}^C \delta_{s_i^{(n)} \geq t'} \leq K \right\}.
\end{align*}

\begin{figure}
    \centering
    \includegraphics[width=.45\textwidth]{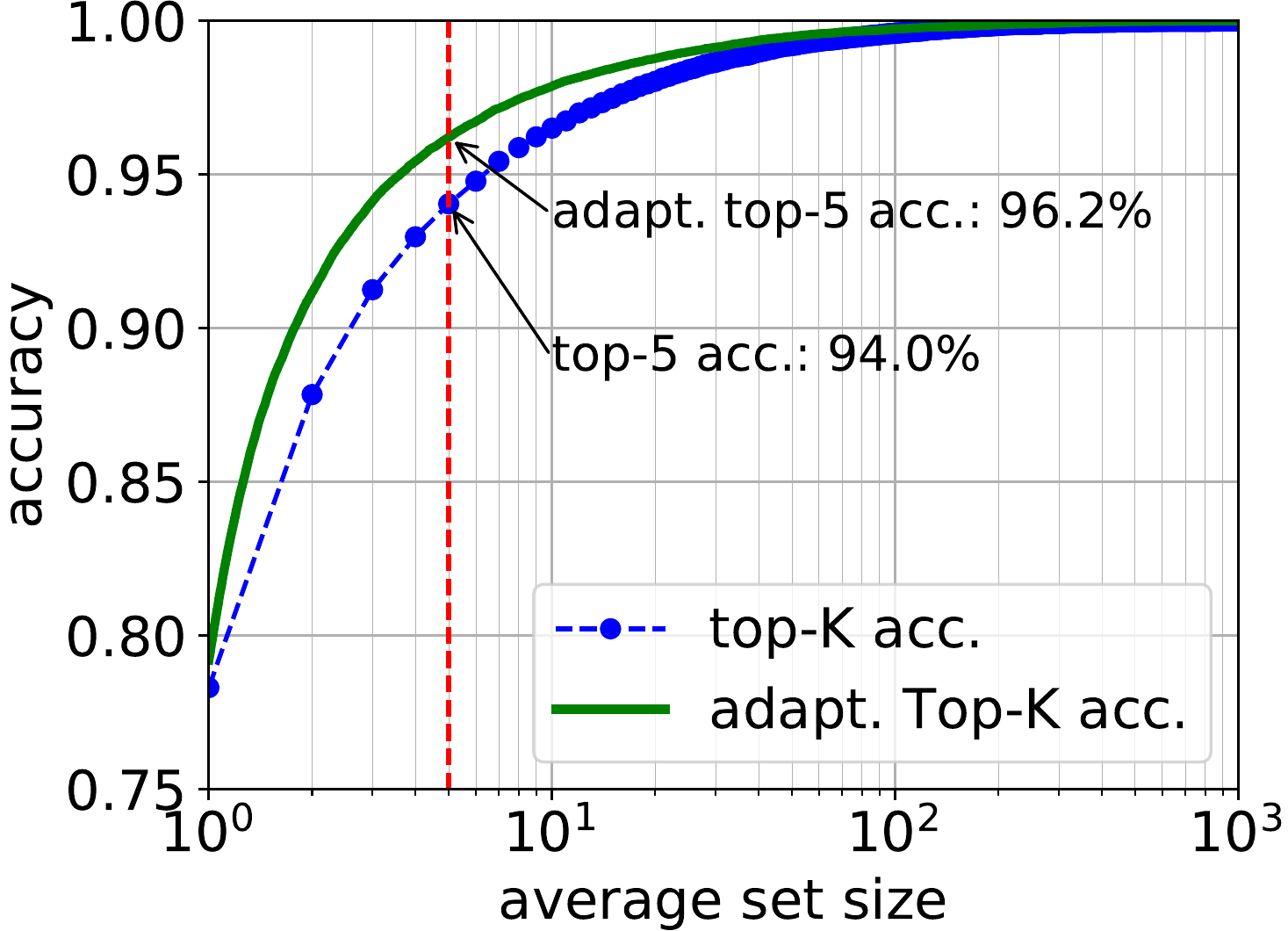}
    \caption{
        Illustration of evaluation criteria on a mock dataset where we are interested in $K=5$.
        The blue and green curve represents, resp., values of top-$K$ accuracy and of adaptive top-$K$ accuracy, for varying values of $K$.
        The red line represents the vertical line $K=5$.
    }
    \label{fig:evaluation-criteria}
\end{figure}

These two evaluation criteria are illustrated in \autoref{fig:evaluation-criteria} on a mock dataset.
The top-$K$ accuracy and its adaptive variant are plotted, resp. in blue and green, for different average numbers of predicted species.
The top-$K$ accuracy is a discrete curve taking values only for integer values of $K$.
On the other hand, the adaptive top-$K$ is a continuous curve.
The curve for the adaptive top-$K$ accuracy is always above the curve for the standard top-$K$ accuracy, since relaxing the hard constraint of predicting a fixed number of species allows more species to be predicted where they are more numerous and fewer where they are scarce, resulting in a higher accuracy.

Instead of looking at the whole curves, we focus on a specific value of $K$ illustrated by the red line.
In particular, we fix $K = 30$ for both evaluation criteria.
This value corresponds to the average observed plant species richness across the inventoried plots of the French botanical data of SOPHY \cite{sophy}.

\section{GeoLifeCLEF 2020 Competition}

The GeoLifeCLEF 2020 competition aims to use this dataset to advance the state-of-the-art in location-based species reccomendation. 
The competition is being hosted on AIcrowd \cite{competitionWeb}. 

To compute the evaluation criterion, we need a score for each class for each data point. 
Due to limitations on the size of the submission files on AIcrowd \cite{competitionWeb}, only the 150 classes with the highest scores should be provided.
We will use these values to compute the metric for the challenge.
However, for further \textit{a posteriori} analysis, we will ask participant to keep and send us the complete predictions for all the classes once the challenge is over.

\section{Acknowledgements}
This work was supported in part by the Microsoft AI for Earth program, an NSF Graduate Research Fellowship (DGE-1745301), and the French National Research Agency under the Investments for the Future Program, referred as ANR-16-CONV-0004 (\#DigitAg).

\bibliographystyle{plain}
\bibliography{references}

\clearpage
\onecolumn
\appendix

\clearpage
\section{Land Cover}

\begin{table*}[htb!]
\small
\centering
\begin{tabular}{| c | c | c | c |} \hline
\textbf{New Code} & \textbf{Original Code} & \textbf{Land Cover Class} & \textbf{Country} \\ \hline
0 & 0 & No Data & Both\\ \hline
1 & 11 & Annual Summer Crops & \multirow{17}{*}{France}\\
2 & 12 & Annual Winter Crops & \\
3 & 31 & Broad-leaved Forests & \\
4 & 32 & Coniferous Forests & \\
5 & 34 & Natural Grasslands & \\
6 & 36 & Woody Moorlands & \\
7 & 41 & Continuous Urban Fabric & \\
8 & 42 & Discontinuous Urban Fabric & \\
9 & 43 & Industrial and Commercial Units & \\
10 & 44 & Road Surfaces & \\
11 & 45 & Bare Rock & \\
12 & 46 & Beaches, Dunes, and Sands & \\
13 & 51 & Water Bodies & \\
14 & 53 & Glacier and Perpetual Snow & \\
15 & 211 & Intensive Grasslands & \\
16 & 221 & Orchards & \\
17 & 222 & Vineyards & \\ \hline 
18 & 11 & Open Water & \multirow{16}{*}{USA}\\
19 & 12 & Perennial Ice/Snow & \\
20 & 21 & Developed, Open Space & \\
21 & 22 & Developed, Low Intensity & \\
22 & 23 & Developed, Medium Intensity & \\
23 & 24 & Developed, High Intensity & \\
24 & 31 & Barren Land (Rock/Sand/Clay) & \\
25 & 41 & Deciduous Forest & \\
26 & 42 & Evergreen Forest & \\
27 & 43 & Mixed Forest & \\
28 & 52 & Shrub/Scrub & \\
29 & 71 & Grassland/Herbaceous & \\
30 & 81 & Pasture/Hay & \\
31 & 82 & Cultivated Crops & \\
32 & 90 & Woody Wetlands & \\
33 & 95 & Emergent Herbaceous Wetlands & \\ \hline
\end{tabular}
\caption{Mapping between land cover classes, their original class codes in the source datasets, and the new class codes used in our dataset.}
\label{tab:landcoverclasses}
\end{table*}

\begin{table*}[htb!]
\small
\centering
\begin{tabular}{| c | c | c |} \hline
\textbf{France} & \textbf{USA} & \textbf{Description}\\ \hline
13 & 18 & Water\\ 
14 & 19 & Ice and Snow\\
11-12 & 24 & Barren Land (Rock/Sand/Clay)\\ 
8-10 & 20-21 & Developed Low Intensity\\ 
7-9 & 22-23 & Developed High Intensity\\
3 & 25-27 & Broad-leaved Forest\\ 
4 & 26 & Coniferous Forest\\ 
6 & 28 & Shrub/Scrub/Woodie moorlands\\
5 & 29 & Natural Grassland/Herbaceous\\
15 & 30 & Pasture/Hay/Intensive grasslands\\ 
1-2-16-17 & 31 & Cultivated Crops\\
- & 32 & Woody Wetlands\\ 
- & 33 & Emergent Herbaceous Wetlands\\
\hline
\end{tabular}
\caption{
    Suggested alignment between the land cover classes from France and the US.
    Note that the class codes are the new codes used in our dataset.
}
\label{tab:landcoveralignment}
\end{table*}

\clearpage
\section{Bioclimatic Data}

\begin{table}[htb!]
    \centering
    \begin{tabular}{|r|l|c|}
    \hline
    \textbf{Name} & \textbf{Description} & \textbf{Resolution} \\
    \hline 
bio\_1 & Annual Mean Temperature & 30 arcsec \\
bio\_2 & Mean Diurnal Range (Mean of monthly (max temp - min temp)) & 30 arcsec \\
bio\_3 & Isothermality (bio\_2/bio\_7) (* 100) & 30 arcsec\\
bio\_4 & Temperature Seasonality (standard deviation *100)& 30 arcsec\\
bio\_5 & Max Temperature of Warmest Month & 30 arcsec\\
bio\_6 & Min Temperature of Coldest Month & 30 arcsec\\
bio\_7 & Temperature Annual Range (bio\_5-bio\_6) & 30 arcsec\\
bio\_8 & Mean Temperature of Wettest Quarter & 30 arcsec\\
bio\_9 & Mean Temperature of Driest Quarter& 30 arcsec\\
bio\_10 & Mean Temperature of Warmest Quarter & 30 arcsec\\
bio\_11 & Mean Temperature of Coldest Quarter& 30 arcsec\\
bio\_12 & Annual Precipitation & 30 arcsec\\
bio\_13 & Precipitation of Wettest Month & 30 arcsec\\
bio\_14 & Precipitation of Driest Month & 30 arcsec \\
bio\_15 & Precipitation Seasonality (Coefficient of Variation)& 30 arcsec\\
bio\_16 & Precipitation of Wettest Quarter & 30 arcsec\\
bio\_17 & Precipitation of Driest Quarter & 30 arcsec\\
bio\_18 & Precipitation of Warmest Quarter & 30 arcsec\\
bio\_19 & Precipitation of Coldest Quarter& 30 arcsec \\

orcdrc & Soil organic carbon content (g/kg at 15cm depth) & 250 m \\
phihox & Ph x 10 in H20 (at 15cm depth) & 250 m \\
cecsol & cation exchange capacity of soil in cmolc/kg 15cm depth  & 250 m\\
bdticm & Absolute depth to bedrock in cm & 250 m\\
clyppt & Clay (0-2 micro meter) mass fraction at 15cm depth & 250 m \\
sltppt & Silt mass fraction at 15cm depth& 250 m \\
sndppt & Sand mass fraction at 15cm depth & 250 m\\
bldfie & Bulk density in kg/m3 at 15cm depth & 250 m\\
    \hline
    \end{tabular}
    \caption{Summary of raster environmental variables provided.}
    \label{tab:my_label}
\end{table}

\section{Data format}
\subsection{Occurrences}
Occurrence data is provided in 5 CSV files located on the AIcrowd page of the challenge \cite{competitionWeb}.
Each country have 2 occurrences files, one including all training occurrences, the other including all testing occurrences.
The training files have 4 columns named ``\texttt{id}'' (IDs of occurrences), ``\texttt{lat}'' (latitude), ``\texttt{lon}'' (longitude) and ``\texttt{species\_id}'' (IDs of corresponding species).
The testing files contains the same columns without the ``\texttt{species\_id}'' (this is what has to be predicted).
The last CSV file named ``\texttt{species\_metadata}'' provides the link between the ``\texttt{species\_id}'', the ``\texttt{GBIF\_species\_id}'' (unique species ID in the GBIF database) and ``\texttt{GBIF\_species\_name}'' (species binomial name in the GBIF taxonomy).

\subsection{Patches}
The patch files are named with their corresponding occurrences ids.
They are formatted in 2 NumPy arrays for each occurrences, one containing 5 layers (red imagery, green imagery, blue imagery, near-infrared imagery, land cover) stored in ``\texttt{uint8}'' and one containing 1 layer (altitude) stored in ``\texttt{int16}''.
To avoid problems due to the number of files, patches are stored in a 2-depth directory structure. For an occurrences with an id \texttt{XXXXABCD} the corresponding patches can be found at \texttt{./CD/AB/XXXXABCD.npy} and \texttt{./CD/AB/XXXXABCD\_alti.npy}.

The patches can be downloaded using the following URLs:
\begin{itemize}
    \item US: \url{https://lilablobssc.blob.core.windows.net/geolifeclef-2020/patches_us_XX.tar.gz}
    \item FR: \url{https://lilablobssc.blob.core.windows.net/geolifeclef-2020/patches_fr_XX.zip}
\end{itemize}
where \texttt{XX} is replaced by \texttt{01}, \texttt{02}, \ldots, \texttt{19}, \texttt{20}. 
The links are also available on the AIcrowd page for the challenge \cite{competitionWeb}. 

\subsection{Rasters}
Rasters are available in a set of folders, one per variable, and each contain two GeoTIFF files: \texttt{XXX\_FR.tif} for France and \texttt{XXX\_USA.tif} for the US. 
They are available on the AIcrowd page for the challenge \cite{competitionWeb}.

\subsection{Code}
To help with data manipulation and setting up experiments, a GitHub repository containing some useful code is also provided. 
See the campaign GitHub page (\url{https://github.com/maximiliense/GLC}) for the code and more details. 
Among other things, this code includes a script for extracting patches from bioclimatic and pedological rasters. 
The function call returns, for each queried point, either the vector of variables values or a 3D array of locally cropped rasters.

\subsection{Size}
\begin{table*}[htb!]
\small
\centering
\begin{tabular}{| c | c |} \hline
\textbf{Data} & \textbf{Size}\\ \hline
Rasters & 2.3 GB \\
Occurrences & 81 MB \\
Patches (France) & 359 GB \\
Patches (US) & 478 GB \\
\hline
\end{tabular}
\caption{Extracted data size on disk.}
\label{tab:datasize}
\end{table*}

\end{document}